\documentclass[conference]{IEEEtran}
\IEEEoverridecommandlockouts
\usepackage{cite}
\usepackage{amsmath,amssymb,amsfonts}
\usepackage{algorithmic}
\usepackage{graphicx,subcaption}
\usepackage{textcomp}
\usepackage{xcolor}
\usepackage{hyperref}
\usepackage{todonotes}
\usepackage[latin9]{inputenc}
\usepackage{amsmath}
\usepackage{graphicx}

\usepackage{lipsum}

\usepackage{xcolor,colortbl}
\definecolor{red}{rgb}{1.00,0.00,0.00}
\definecolor{blue}{rgb}{0.00,0.00,1.00}

\newcommand{\cblue}[1] {\textcolor{blue}{#1}}

\usepackage{url}
\def\BibTeX{{\rm B\kern-.05em{\sc i\kern-.025em b}\kern-.08em
    T\kern-.1667em\lower.7ex\hbox{E}\kern-.125emX}}

\makeatletter
\def\ps@IEEEtitlepagestyle{%
  \def\@oddfoot{\mycopyrightnotice}%
  \def\@evenfoot{}%
}
\def\mycopyrightnotice{%
  {\footnotesize 978-1-6654-3198-9/21/\$31.00\ \textcopyright2021 IEEE\hfill}
  \gdef\mycopyrightnotice{}
}

\begin{document}
\title{{Few-Shot Visual Grounding \\for Natural Human-Robot Interaction}}
\author{\IEEEauthorblockN{{Giorgos Tziafas}}
\IEEEauthorblockA{
\textit{Department of Artificial Intelligence} \\
\textit{University of Groningen}  \\
\texttt{g.tziafas@student.rug.nl}} 
\and
\IEEEauthorblockN{{Hamidreza Kasaei}}
\IEEEauthorblockA{
\textit{Department of Artificial Intelligence} \\
\textit{University of Groningen}  \\
\texttt{hamidreza.kasaei@rug.nl}} 
}

\maketitle
\begin{abstract}
Natural Human-Robot Interaction (HRI) is one of the key components for service robots to be able to work in human-centric environments. In such dynamic environments, the robot needs to understand the intention of the user to accomplish a task successfully. Towards addressing this point, we propose a software architecture that segments a target object from a crowded scene, indicated verbally by a human user. At the core of our system, we employ a multi-modal deep neural network for visual grounding. Unlike most grounding methods that tackle the challenge using pre-trained object detectors via a two-stepped process, we develop a single stage zero-shot model that is able to provide predictions in unseen data. We evaluate the performance of the proposed model on real RGB-D data collected from public scene datasets. Experimental results showed that the proposed model performs well in terms of accuracy and speed, while showcasing robustness to variation in the natural language input.
\end{abstract}
\section{Introduction}
\label{intro}
Humans have the cognitive capacity to process multi-modal data (e.g. vision, language) and make cross-references between parts of the two modalities in real-time effortlessly. 
They are also very capable of identifying such cross-references in degenerate cases where one modality suffers from noise. However, this is not the case in robotics, where most commonly the visual perception and HRI modules are treated separately. 
In this regime, when verbal input is given to the robot (e.g. in the context of grasping: ``Grasp the \textit{mug}'') the referring phrases that correspond to objects to be segmented (aka \textit{grounded},  e.g. ``\textit{Mug}'') must be predefined explicitly and hard-coded in the agents behaviour. 
As a result, the agent is unable to comprehend variants of the predefined object category from the verbal input as it is often the case in real-world scenarios, where objects might be referred by their visual attributes or spatial relation to another object (e.g. ``Grasp the \textit{red mug}'' or ``Grasp the \textit{mug} that is \textit{next to} the \textit{laptop}'' in the case of multiple mug objects within view). 
A more \textit{natural} HRI setting would require a bridging between the two modules, allowing the model to interpret ambiguous verbal references and generalize over unseen object categories. 
This paper proposes such a bridging that tackles the problem by employing an end-to-end multi-modal deep learning model, able to make predictions for image-phrase pairs that were never seen in the training data.

\begin{figure}[!t]
    \centering
    \resizebox{\linewidth}{!}{
    \begin{tabular}{ccc}
         \includegraphics[width=.3\linewidth]{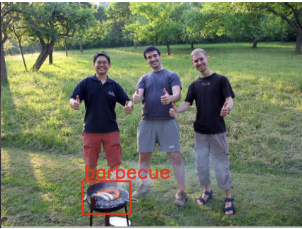}&
         \includegraphics[width=.3\linewidth]{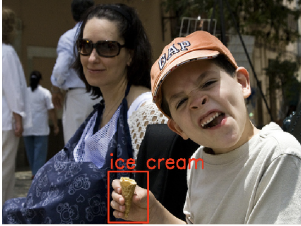}&
         \includegraphics[width=.3\linewidth]{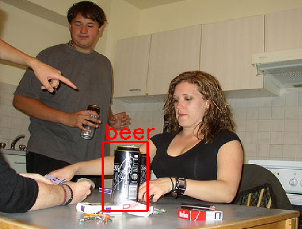}\\
         \scriptsize{\texttt{barbecue}} & \scriptsize{\texttt{ice cream}} & \scriptsize{\texttt{beer}} \\
         \includegraphics[width=.3\linewidth]{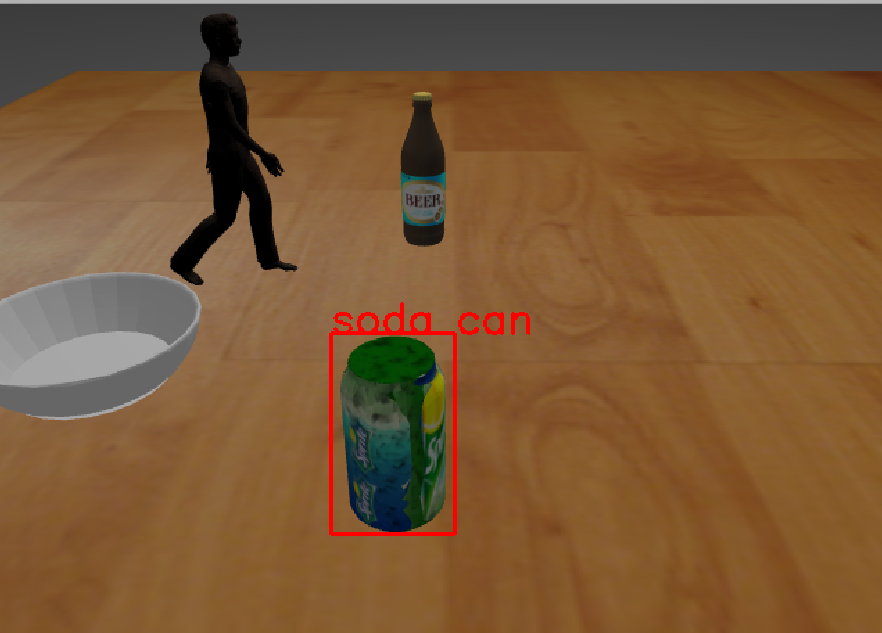}&
         \includegraphics[width=.3\linewidth]{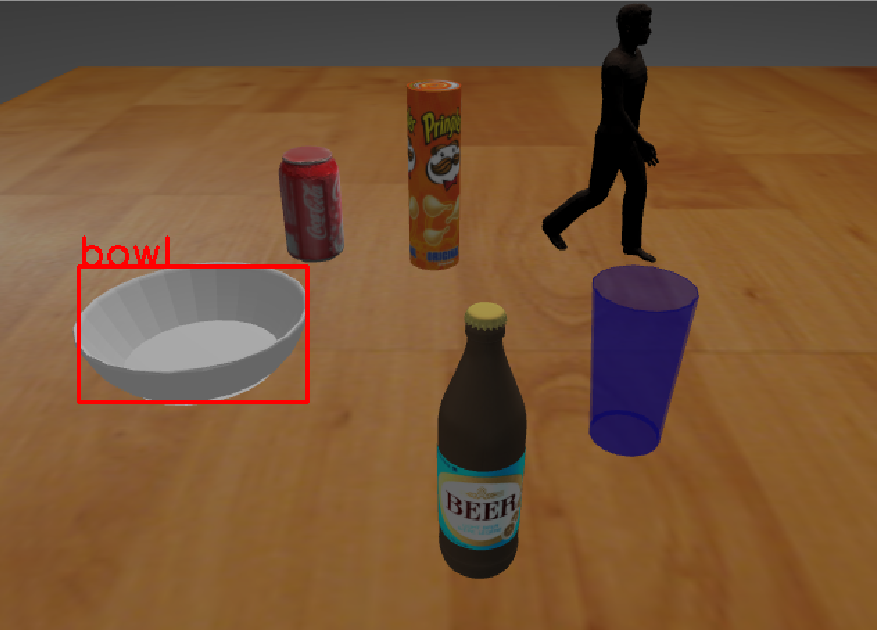}&
         \includegraphics[width=.3\linewidth]{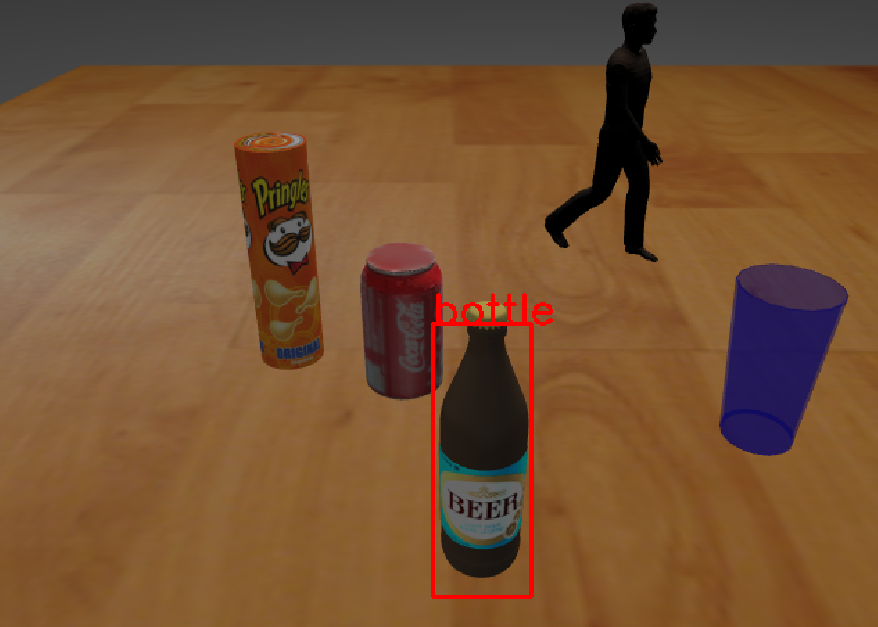}\\
       \scriptsize{\texttt{soda can}} &\scriptsize{ \texttt{bowl}} & \scriptsize{\texttt{bottle}} \\
    \end{tabular}}
    \caption{Examples of grounding predictions made by our model for real (top) and simulated (bottom) scenes along with the corresponding input phrase queries.}
    \label{fig:examples}
\end{figure}

The task at hand is described in literature as \textit{single-query visual grounding} or \textit{referring expression comprehension}. In this task, given an input image and a natural language phrase referring to a specific entity, the model has to localize the entity inside the frame (see Fig.~\ref{fig:examples}). Several research has been done to address this task, both solely and as a proxy task for \textit{visual question-answering} (VQA) and \textit{visual reasoning} downstream applications. Section~\ref{sota} provides a brief overview of such research directions.
In this work, we aim to design a grounding agent that can be utilized in a practical HRI scenario, such as grasping household objects based on verbal commands given by a human supervisor. To be efficient for this application, our agent needs to be able to infer groundings at high speed. Also, when a broad variation of object categories, as well as robustness to the phrasing of the referring expressions describing them, is desired, the system should be able to provide predictions for object - phrase instances that it has never encountered during training. These considerations guide our choice of deep learning model and the architecture of the implemented system, both of which are described in Section~\ref{methods}. 

We explore the robustness of the proposed approach to variations of the natural language input by probing it in test time with extensive captions of multiple levels of reference, namely: a) multiple instances of the same entity  (plural) or b) a specific instance of an entity appeared multiple times in the scene by referring to its visual attribute (color) or some other property (e.g. gender in the case of humans). Additionally, as most benchmarks for visual grounding evaluate on domain-agnostic datasets, in this work we opt to explore the grounding models performance on an HRI-oriented data domain, by evaluating its zero/few-shot performance in two tiny-scale datasets of image-query pairs respecting the assumed HRI scenario mentioned above. The image samples are collected from sub-sets of two RGB-D datasets, namely: \textit{RGBD-Scenes} \cite{rgbd-scenes}, used most popularly for 3D vision learning and the \textit{Objects Cluttered Indoors Dataset (OCID)} \cite{ocid}, used for manipulation planning. Details about the evaluation data and some results are presented in section~\ref{Results}. Conclusion and some interesting potential future research directions are given in section~\ref{Conclusion}.

\section{Related work}
\label{sota}
Visual grounding architectures typically use object detectors such as Faster-RCNN \cite{faster-r-cnn} as a pre-processing step to extract bounding box proposals and corresponding object features, thus limiting the categories that the model is able to learn to the ones pre-specified by the pre-trained detector \cite{sadhu2019zero}. Refined versions use cross-modal self-attention mechanisms to capture long range dependencies in the two modalities \cite{cross-modal-self-attn}. More recent approaches extract scene graphs from images-text and tackle grounding as a structure prediction task where visual and textual {Graph Neural Networks (GNNs)} are used to contextualize each modalitys representations and a graph similarity metric is introduced to prune the two graphs appropriately \cite{context-graphs}. 

Another family of state-of-the-art models is that of pre-trained representation learners, such as {VisualBERT}~\cite{visual-bert} and {ViLBERT}~\cite{vilbert}. Such models are very parameter-heavy, trained on massive-scale datasets with self-supervised objectives, such as language modeling, masked region modeling, word-region alignment and image-text matching. They often also apply a \textit{multi-task setting} \cite{12in1}, in which supervision from multiple tasks-domains is injected during self-supervised pre-training. Single-task performance is then benefited from attempting to solve all tasks at once. On the downside, the enormous scale of these models grant them potentially inefficient for real-time application.

Several robotics-grounded works attempt to explicitly model natural language input for HRI applications, either as a purely textual or as an end-to-end multi-modal system. 
In some works, the system maps input natural language instructions into action sequences for task handling~\cite{tellmedave, hr-dialogue}. 
Special care is given so that the model handles variations of natural language input and is able to generalize over unseen environment and task settings. 
Paul et. al.~\cite{spatial_concepts} implemented a probabilistic model that parses natural language input into object/region proposals as well as motion constraints by incorporating notions of abstract spatial concepts. Shridhar et. al.~\cite{hri_} presented an HRI-applied multi-modal system for visual grounding, where the model was able to handle variations of natural language input (for situations of multiple object instances) by interacting with the user through an ambiguity-resolution question, generated by an auto-regressive RNN which is trained jointly with the grounding network. One drawback of their work is that they employed a two-stage approach for providing groundings, intensifying computational requirements in inference time. In this work, we attempt to address this limitation by employing a one-stage zero-shot model.

\section{System Architecture}
\label{methods}
In this section we present the visual grounding architecture that is developed as well as the technical details of the overall architecture of the system.

\subsection{Learning Model}
\label{model}
The deep learning model that is chosen as a baseline for our implementation is the \textit{Zero-Shot Grounding (ZSG)} network~\cite{sadhu2019zero}. Similar to single-shot architectures employed in object localization, such as \cite{ssd}, the proposed model generates bounding box proposals that refer to the input query based entirely on the size of the input image. As a result, an end-to-end trainable image encoder for capturing image representations can replace pre-trained object descriptors. For each image-query input pair, the model generates a set of bounding box proposals $B = \{b_1,b_2,...,b_N\}$ and outputs the best candidate box $b_i$, as well as four regression parameters $\{x_1,y_1,x_2,y_2\}$ that correspond to the updated top-left and bottom-right box coordinates in the input image frame. Since this is an end-to-end architecture, the visual features extracted by this model are independent of the trained object intra-class variance. Therefore, the proposed method departs from the limitations posed by other pre-trained visual grounding systems, and is suitable for real-time applications due to its single stage design and computational efficiency, especially during inference.

The model consists of five main components, including: (\textit{1}) a language module that encodes the query phrase into a continuous vector space, (\textit{2}) a visual module that extracts multiple image feature maps, (\textit{3}) an anchor generator for proposing multiple scale bounding boxes, (\textit{4}) a multi-modal fusion scheme for injecting all features into a single representation and a linear layer that predicts the most likely box proposal, (\textit{5}) as well as the array of regression parameters for its fixed coordinates. For coherence purposes, we provide a brief overview of these modules in the following sub-sections~\cite{sadhu2019zero}. 

\subsubsection{{Language module}} This module consists of an embedding layer followed by a recurrent neural encoder for encoding the input query phrase. The embedding layer is responsible for mapping each word $W_i$ in our vocabulary to a dense vector $\vec{w_i} \in \mathbb{R}^{d_w}$. The encoder is a uni-layered bi-directional LSTM architecture \cite{bi-lstm} that processes sequentially the entire input word vector sequence $\{\vec{w_i}\}, i=1,...,T,$ in both directions (start to end and vice-versa) and at each step outputs a hidden state vector $\vec{h_i} \in \mathbb{R}^{2d_w}$ that is informed by the context of the entire phrase in both directions at this point. The encoding that we use for representing the entire phrase is the hidden state vector $\vec{h_T}$ produced in the last time step.

Alternatively, we implement phrase contextualization using a single \textit{Transformer Encoder} layer~\cite{attention} and aggregate the hidden representations $\mathbf{H} \in \mathbb{R}^{T \times d_w}$ for the entire sequence with average pooling.

\subsubsection{{Visual module}} This module consists of a deep \textit{Convolutional Neural Network} (CNN) that learns how to represent the input 2D image into a dense feature map. This is a standard CNN architecture used for object recognition without the linear layers that are used for cross-entropy classification. In our version, $K$ feature maps $\mathbf{V_j} \in \mathbb{R}^{d_v \times d_v} , j=1,...,K,$ are extracted at different resolutions. The model used for encoding is a ResNet-50 \cite{resnet50}, augmented into the \textit{Feature Pyramid Network} (FPN) architecture \cite{FPN} for extracting multi-scale hierarchical feature maps. The feature maps are normalised across the channel dimension.

\subsubsection{{Anchor generation}} The first step is to form a grid. For each cell of the grid,  anchors of different shapes are proposed. The convention of the grid is (-1,-1) to (1,1), similar to the one defined for bounding boxes. Anchors are defined in terms of ratios and scales. The ratio is the ratio of the height of an anchor box to its width. The scale is used to calculate the unit of the height and width of each anchor box. Given a scale ($s$) and a ratio ($r$), the aspect is [$s\sqrt{r}$, $ \frac{s}{\sqrt{r}}$]. These aspects are then resized according to the size of each cell of the grid. The center of each anchor box is considered the centre of the grid cell.

\subsubsection{{Multi-modal fusion}} The language feature vector extracted by the neural encoder is expanded to fit the dimensions of the $K$ extracted visual feature maps and it is concatenated along the channel dimension of each feature map, so that $\mathbf{H_T} \in \mathbb{R}^{d_v \times d_w}, \; \mathbf{H_T}=\begin{pmatrix}
 \vec{h_t}& \vec{h_t} & ... & \vec{h_t}
\end{pmatrix}^T$. The generated anchor box centres are also appended at each cell of the feature maps. The resulting multi-modal feature representations $\mathbf{M_j} \in \mathbb{R}^{d_m \times d_m}$ are then given for $j=1,...,K$ by:

\begin{equation*}
   \mathbf{ M_j}(x,y) = \mathbf{V_j}(x,y) \; ; \; \mathbf{H_T}(x,y) \; ; \; \frac{c_x}{W} \; ; \; \frac{c_y}{H}
\end{equation*}

\noindent where $;$ operator denotes the concatenation, $c_x,$ and $c_y$ represent the center locations of the normalised feature maps at each $(x,y)$ location. The initial size of the input image is represented by $W$, and $H$. The scaling operation is performed in order to aid location-based grounding, when input query phrases contain location information, providing functionality for making spatial references between recognised objects in the input image frame.

\subsubsection{Anchor matching} Each generated anchor proposal of different size is matched to every cell of the produced multi-modal feature maps. For each box, a linear layer maps the multi-modal features into a 5-dimensional vector containing the prediction score (confidence) and the regression box parameters that update the coordinates to bound the referenced object tightly.  

\subsection{Online Object Segmentation}
\label{system}
\begin{figure}[]
\centering
\includegraphics[width=1\linewidth]{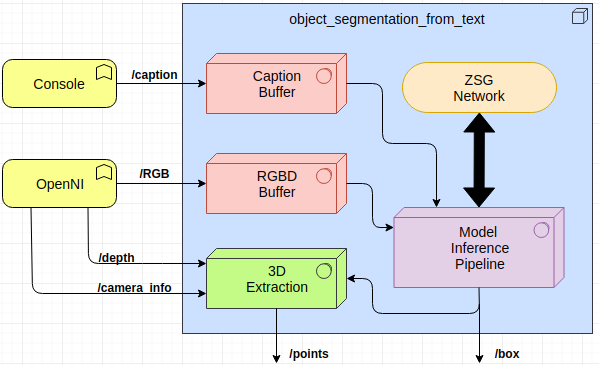}
\caption{A schematic of the implemented \textit{ROS} system. The inference pipeline is implemented through an action client that given an input RGB frame and a caption sends a request to our network and outputs the coordinates of the most probable bounding box. A 3D Extraction node extracts a \textit{PointCloud} by combining the RGB box with the corresponding depth box from the registered depth frame.}
\label{fig:system}    
\end{figure}

In this section, we describe the design of the implemented software architecture, shown in Fig.~\ref{fig:system}. 
Our system receives a stream of RGB-D topics produced real-time from a depth sensor along with a natural language caption given by a human user and produces an RGB image with the predicted bounding box drawn into the frame, as well as a \textit{PointCloud} containing 3D information of the segmented object. 
The images on the RGB-D buffer are refreshed with each new pair and only when a caption is given the last entry is grabbed and used as input to the network.

For real-time visual grounding, we have implemented an action server that functions as the communication side between the  sensor data and the implemented ZSG network. At its spawn, the server loads all the necessary utilities (word embeddings etc.) and loads a pre-trained instance of the ZSG network. When a caption is buffered, a client request for a grounding inference is published. The action server appropriately pre-processes the input raw RGB frame in order to enhance it's contrast, by applying \textit{Contrast Limited Adaptive Histogram Equalisation} (CLAHE) \cite{CLAHE} in the \textit{YUV} colourspace. The query phrase is embedded and the resulting image-query pair is passed as input to the network for a single forward pass. After inference, some post-processing steps for drawing the predicted box and segmenting the foreground for 3D extraction are applied and the resulting frames are published. The server responses are executed in $>30$Hz, even without utilizing GPU-enabled acceleration like during training (tested in an \textit{Intel Xeon E5-2680 v3 @ 2.50GHz} CPU). 
\section{Experimental Results}
\label{Results}

\subsection{Training}
\label{data}

We trained our model using two benchmark visual grounding datasets.
The first training dataset {was} \textit{Flickr30k Entities} \cite{flickrentitiesijcv}, comprised of $30\,000$ annotated images associated with five sentences, each having multiple entity queries referring to the image with an average of $3.6$ queries per sentence. As a result, each sentence {has been} split to provide more grounding samples for the respective image. The annotations are bounding boxes containing the category labels of the grounded entity.
The second dataset was the \textit{RefClef} strain of the \textit{ReferIt} dataset \cite{referit}, itself being a subset of \textit{Imageclef} \cite{imageclef}, enumerating approximately $20\,000$ images with a total of $85k$ single-query caption sentences. For both datasets, we used the same training-validation-test splits as in \cite{splits}.

We considered two options for embedding input language semantics into word vectors for English, namely: (\textit{i}) pre-trained \textit{GloVe} embeddings \cite{GloVe} and (\textit{ii}) the last hidden layers representations of a pre-trained BERT-BASE \cite{bert}, followed by a trainable linear layer down-scaling the vectors to desired dimensionality. For extracting visual features we utilized a \textit{RetinaNet} \cite{focal_loss} network with a \textit{ResNet-50} \cite{resnet50} backbone. The image samples were resized to $300\times300$ and the word embeddings size as well as the hidden size of the bi-LSTM contextualizing the input phrase was set to $300$, so that both feature vectors could be appended to fit the dimensionality of the multi-modal feature maps. The input query phrases were padded to a maximum length of $50$ words per phrase. Following \cite{sadhu2019zero}, a total of $9$ candidate anchor proposals of different sizes is generated. 

For formulating the supervision signal as well as quantifying the evaluation performance, we utilised the \textit{Intersection over Union} (IoU) metric, calculated as the total overlapping area between the proposed box and the ground truth box. Following the original implementation, we used an \textit{IoU} threshold of 0.5, meaning that only proposals that fit the ground truth box's area over 50\% are considered candidate:
\begin{equation*}
    g_{b_i} \doteq  1 \cdot [ IoU(b_i,gt)\geq 0.5] \; + \; 0 \cdot [IoU(b_i,gt)<0.5],
\end{equation*}
\begin{equation*}
    G \doteq \{b_i \; | \; g_{b_i}=1\}
\end{equation*}
\noindent where $B=\{b_i, \; i=1,...,9\}$ represents the set of all proposals and $gt$ states the ground truth box. The set $G$ is the collection of all candidate proposals $(g_{b_1}=1)$.
The training loss is then calculated as the sum of the focal loss $L_F$ described in \cite{focal_loss} (applied for deciding the prediction score of the i-th proposal $p_{b_i}$ for its binary classification as foreground that possibly contains the query vs. background), with parameters $\alpha=0.25, \; \gamma=2$ and the smooth-L1 loss $L_S$ \cite{smooth_L1_loss} (used for regressing the parameters $r_{b_{i}}$ of a tighter matching bounding box for the most probable proposal):

\begin{equation*}
    L = \frac{1}{\left |G  \right |}\sum_{i=1}^{\left |B  \right |}L_F(p_{b_i}, g_{b_i}) + \frac{1}{\left |G  \right |}\sum_{i=1}^{\left |B  \right |}g_{b_i} \cdot L_S(r_{b_i},gt)
\end{equation*}
The reported accuracies are measured as the total average of correctly classified samples with an \textit{IoU} $>0.5$. 
The end-to-end model was trained for a total of $12$ hours for each dataset (around $7-8$ epochs), where the validation loss had already saturated. We utilised the Adam optimizer \cite{Adam} with a learning rate of $10^{-4}$ and a weight decay of $10^{-4}$ for regularisation. The model was trained on a \textit{NVidia Tesla v100} GPU node for parallel processing of batches of $128$ image-query pairs. All code is written in \textit{Python} using the \href{https://pytorch.org/}{\textit{PyTorch}} deep learning framework. Training results compared with several baselines are reported in Table~\ref{tab:zsg}.
\begin{table}[]
    \centering
	\resizebox{0.75\linewidth}{!}{
    \begin{tabular}{|c|c|c| }
    \hline \textbf{Method} & \textbf{Flickr30k}& \textbf{RefClef} \\
    \hline 
    \hline 
    {QRC$^{\star}$ \cite{iou_ref}} & 60.21\% & 44.10\% \\
    \hline 
    {CITE$^{\star}$ \cite{CITE}} & 61.89\% & 34.13\%\\
    \hline 
    {QRG$^{\star}$ \cite{iou_ref}} & 60.1\% & -\\ 
    \hline 
    {ZSG \cite{sadhu2019zero}} & 63.39\% & \textbf{58.64}\% \\ 
    \hline 
    {VilBERT \cite{12in1}} & \textbf{64.61\%} & - \\
    \hline 
    \hline
    {ZSG (Glove-LSTM)} & 62.73\% &  53.44\%\\
    \hline 
    {ZSG (Glove-TRM)} & 61.74\% &  50.41\%\\
    \hline 
    {ZSG (BERT-LSTM)} & \textbf{63.09\%} &  \textbf{54.21}\%\\
    \hline 
    {ZSG (BERT-TRM)} & 62.13\% &  52.12\%\\
    \hline
    \end{tabular}}
    \caption{ Best top-1 accuracies @{IoU = $0.5$} of the implemented model variants after 12 hours of training in both datasets. Results compared to original implementation and other state-of-the-art methods. Methods with $\star$ further fine-tune their network on the entities of the {Flickr30k} dataset.}
    \label{tab:zsg}
\end{table}

Even though the pre-trained BERT language module provides marginally better embeddings for visual grounding, its massive scale sets a huge computational bottleneck to the system. Therefore, we implement our agent based on the pre-trained Glove model.

\subsection{Evaluation}
\begin{figure}
    \begin{tabular}{c}
    \includegraphics[width=0.95\linewidth]{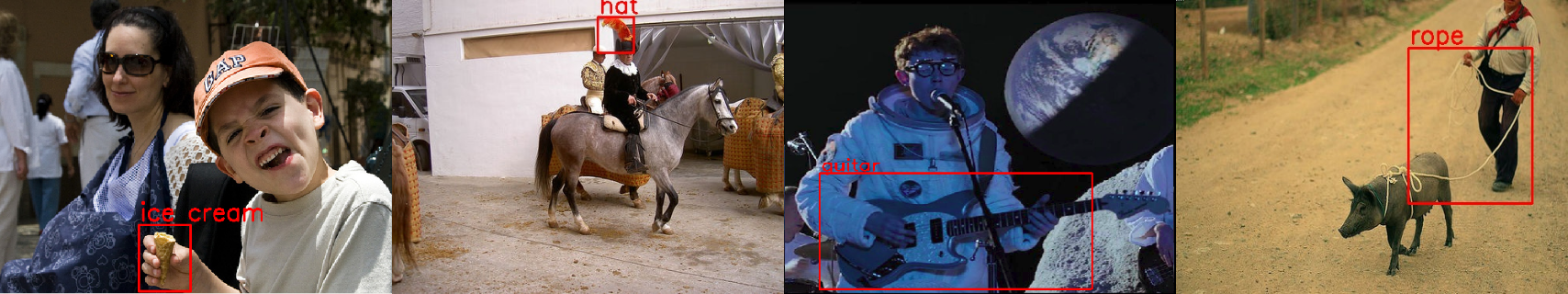}\\
    \scriptsize{(a) Correctly segmented objects}\\
    \includegraphics[width=0.95\linewidth]{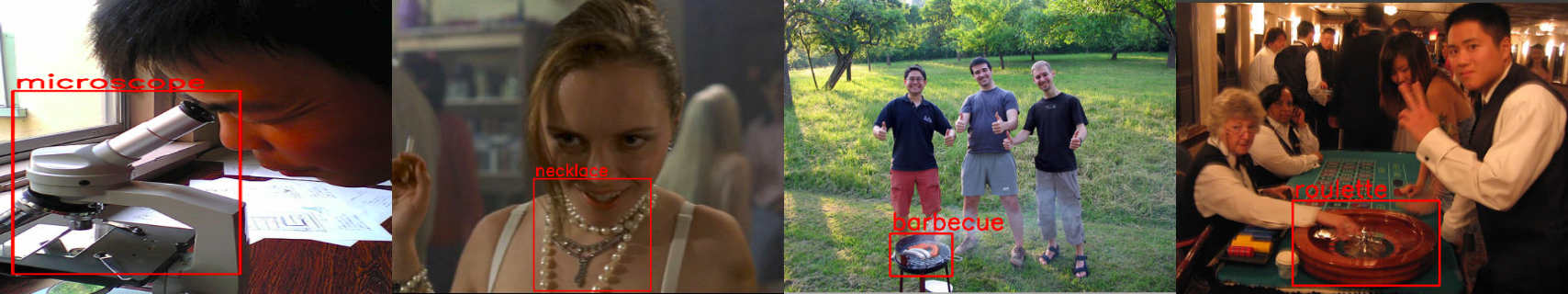}\\
    \scriptsize{(b) Correctly segmented objects with low number of occurrences in training data}\\
    \includegraphics[width=0.95\linewidth]{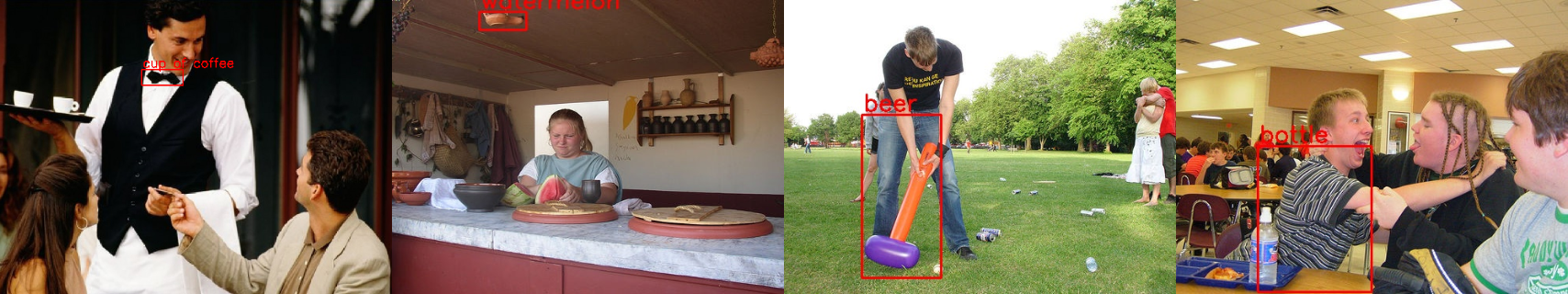}\\
    \scriptsize {(c) Incorrectly segmented objects}\\
    \end{tabular}
    \caption{Several examples of correct and some of incorrect bounding box predictions for image-query pairs. Queries are from left to right: (a) ice cream, hat, guitar, rope, (b) microscope, necklace, barbecue, roulette, (c) cup of coffee, watermelon, beer, bottle.}
    \vspace{-2mm}
    \label{fig:FirstItem}
\end{figure}

In order to evaluate the performance of the proposed system, we conducted multiple experiments, aiming to test not only the predictive accuracy but also the generalisation potential of the proposed model in variants of the standard input query phrases. In this vein, we performed online interactive sessions with our implemented agent, during which humans provide captions for arbitrary images and evaluate the system's predicted segmentation qualitatively. Different depicted objects are queried in free linguistic fashion, also including ambiguous query phrases (plural, referring to a visual cue etc.) that challenge our system in regards to the natural HRI element that we strive for. Figure~\ref{fig:FirstItem} demonstrates some examples of such experiments and a live recorded demo of a session is available online at \href{https://youtu.be/kgQgaghf71o}{\cblue{\texttt{\small{https://youtu.be/kgQgaghf71o}}}} 

Real-time sensory data streamed from a depth sensor, often suffer from various forms of noise compared to the digitally pre-processed images of the training data. To investigate the performance of the proposed system in sensor data, we collected two small-scale datasets of pre-recorded RGB-D images. The scenes included in our samples respect the assumed HRI scenario, i.e., they always comprise of a collection of objects in arbitrary spatial arrangements on top of a planar surface in clear front view of the RGB-D sensor.  

\begin{table}[!b]
    \centering
	\resizebox{\linewidth}{!}{
    \begin{tabular}{|c|c| c| c| c|}
     \hline\textbf{Dataset} & \textbf{\#Img.} & \textbf{avg.QPI} & \textbf{\#Samp.} & \textbf{Mult.Inst. (\%) }  \\
     \hline 
     {RGBD-Scenes} & 67&  3.13& 210 & 17.97\%\\ 
     \hline 
     {OCID} & 18& 2.94& 53& 22.22\%\\
     \hline 
    \end{tabular}}
    \caption{Total number of image samples. average number of query phrases per image, total number of samples (query-phrase pairs) and percentage of samples that include multiple instances of the same object category.}
    \label{tab:stats}
\end{table}
The first dataset is a sub-set of the \textit{RGBD-Scenes} dataset \cite{rgbd-scenes}, which has a total of $67$ RGB-D pairs depicting $22$ different typical household object instances from $6$ different categories (bowl, cap, cereal box, coffee mug, flashlight, soda can) in $8$ different scenes (including multiple desk settings, kitchen, meeting room, small and large table). Bounding box annotations come packed with the dataset. We appropriately sub-sampled the raw training data to include scenes that only contain different object categories or the same objects but viewed from a different angle, resulting in different spatial relationships between them. The second dataset is a curated sub-set of OCID \cite{ocid}, containing a total of $89$ different object categories in two scene settings (table and floor) and two different camera placements (front and top view). As this dataset is originally intended for learning motion planning of a robotic actuator in scenes with highly cluttered objects, we filter the raw data to include only samples with none to minimal cluttering. Statistics about the collected RGB-D datasets are described in Table~\ref{tab:stats}.

\begin{figure}[!t]
    \begin{tabular}{c}
    \includegraphics[width=0.95\linewidth]{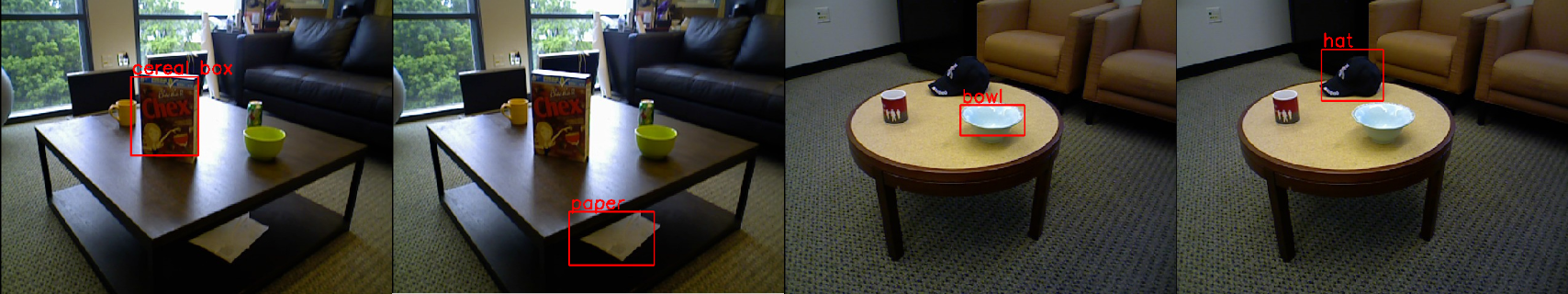}\\
    \scriptsize{(a) Correctly segmented objects from {RGBD-Scenes}} \\
    \includegraphics[width=0.95\linewidth]{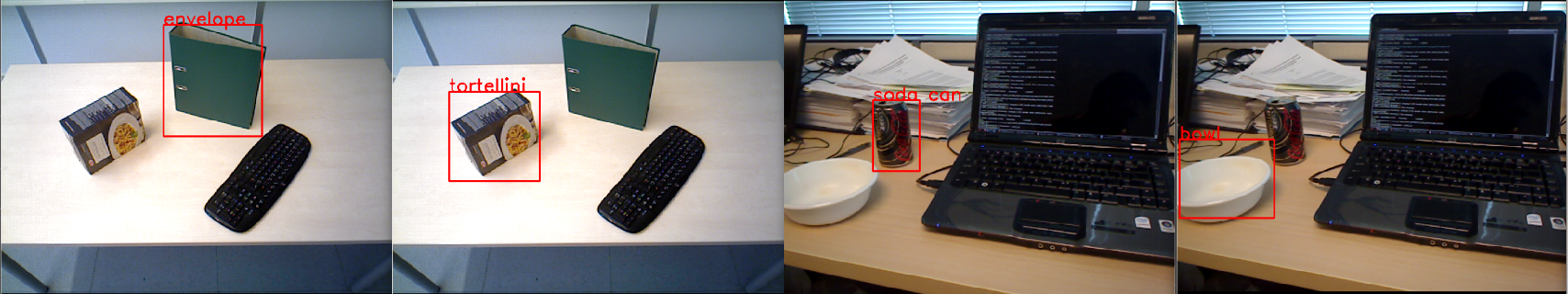}\\
    \scriptsize{(b) Correctly segmented objects with relative cluttered environments}\\
  \includegraphics[width=0.95\linewidth]{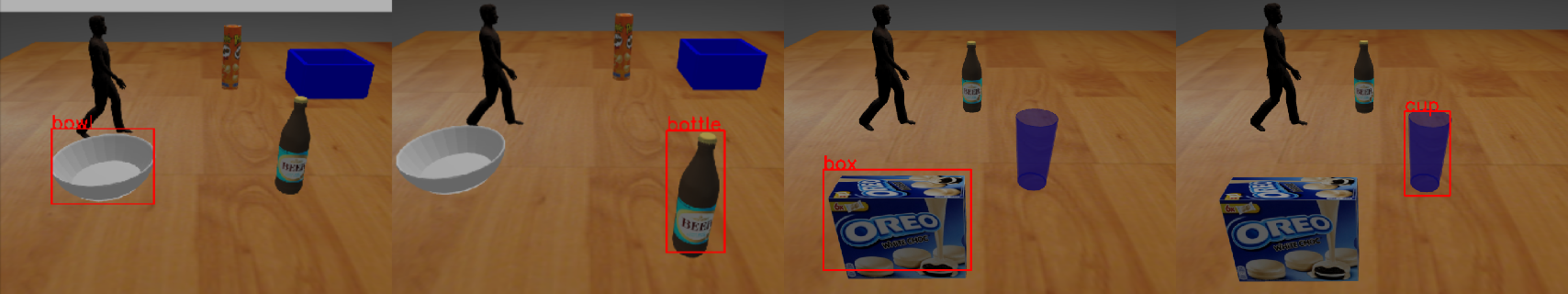}\\
    \scriptsize {(c) Correctly segmented objects in synthetic data simulated in {Gazebo}}\\
    \end{tabular}
    \caption{Several examples of bounding box predictions for real-time RGB-D sensory data from real/synthetic scenes. Queries are from left to right: (a) cereal box, paper, bowl, hat, (b) envelope, tortellini, soda can, bowl, (c) bowl, bottle, box, cup.}
    \label{fig:SecondItem}
    \vspace{-2mm}
\end{figure}

We again qualitatively evaluated and reported accuracy scores representing the zero-shot performance of our system in both domains. We performed preliminary experiments for evaluating the \textit{domain adaptation} potential of our system, where we fine-tuned our model in one RGB-D-scene (e.g. desk) and evaluated in another (e.g. kitchen). By this we aimed to simulate the fine-tuning step that a potential user of our system can employ after gathering minimal samples from the specific object catalogue / environment they wish their agent to operate. Fine-tuning had been implemented identically to the training process with the exception of using a batch size of $32$. The results are presented in Table~\ref{tab:res} and examples of real-time inferences are shown in Figure~\ref{fig:SecondItem}.

\begin{table}[!b]
    \centering
	\resizebox{\linewidth}{!}{
    \begin{tabular}{|c|c|c|c|c|}
     \hline & \textbf{OCID} & \textbf{RGB-D} & \textbf{RGB-D} & \textbf{RGB-D}  \\
     \textbf{Model} & \textbf{(full)} & \textbf{(full)} &  \textbf{(kitchen)} & \textbf{(tables)}  \\ 
     \hline 
     {ZSG-Flickr30k} & 32.13\%&  34.19\%& 32.50\% & 34.13\%\\ 
     \hline 
     {ZSG-RefClef} & 31.37\%& 33.50\%& 36.35\% & 34.13\%\\
      \hline
      {+desk} & -& -& 49.49\%& 50.15\%\\
      \hline 
      {+desk+meeting} & -& -& 55.90\%& 58.20\%\\
       \hline
    \end{tabular}}
    \caption{Evaluation results for both pre-trained models in HRI-specific data, collected from sub-sets of the \textit{RGBD-scenes} and \textit{OCID} datasets. The $+$ rows denote few-shotting results of the ZSG-Flickr30k model in specific scenes after fine-tuning in similar examples from different scenes}
    \label{tab:res}
\end{table}

\begin{figure}[]
\begin{tabular}{c}
    \includegraphics[width=0.95\linewidth]{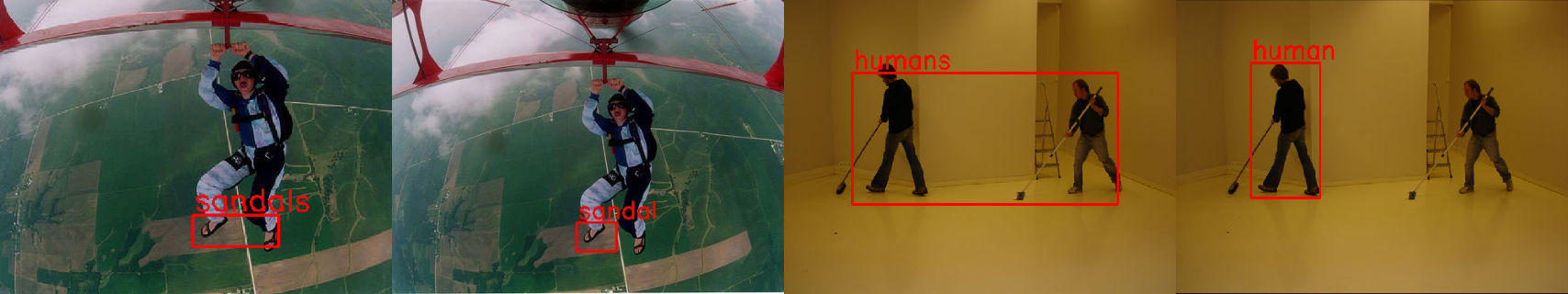}\\
    \scriptsize{(a) Comprehension of plural} \\
    \includegraphics[width=0.95\linewidth]{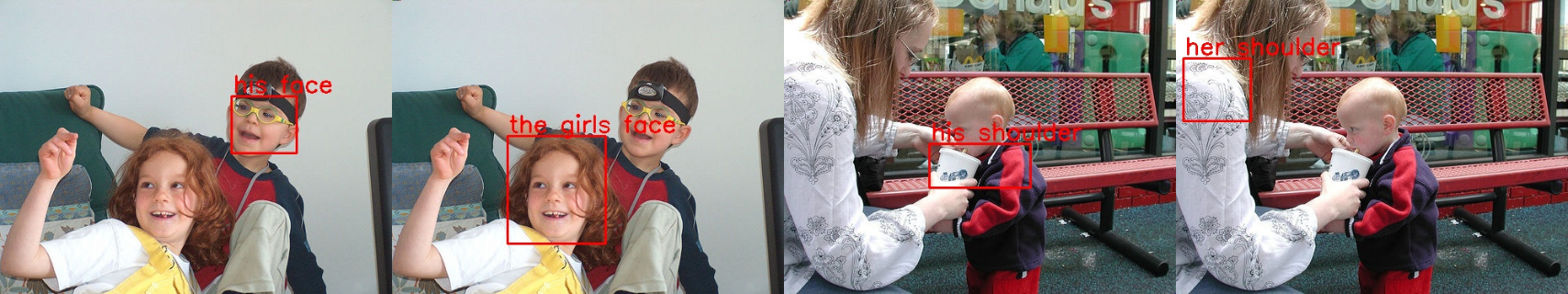}\\
    \scriptsize{(b) Comprehension of pronouns}\\
    \includegraphics[width=0.95\linewidth]{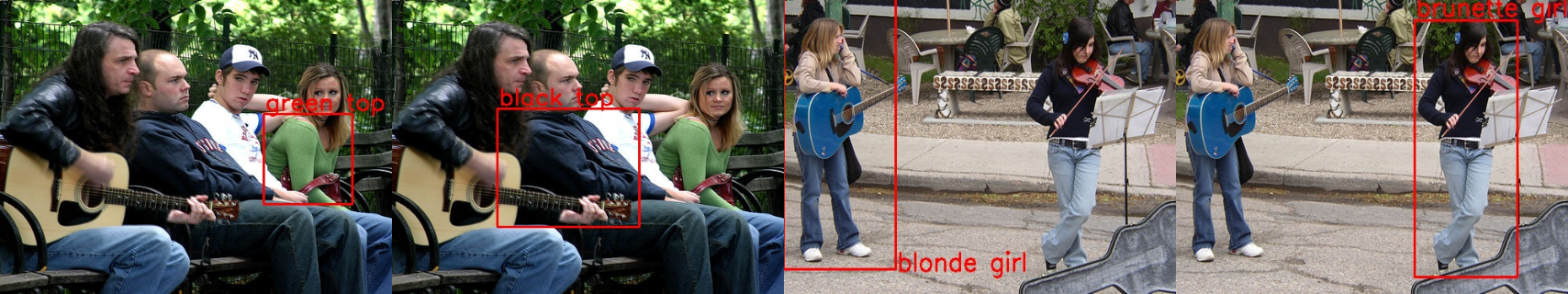}\\
    \scriptsize {(c) Comprehension of visual cues (colour)}\\
    \end{tabular}
    \caption{The system is capable of generalising over variants of the natural language input (plural, pronouns, visual cues). Queries are from left to right: (a) sandal(s), human(s) (b) his/the girls face, his/her shoulder (c) green/black top, blonde/brunette girl.}
    \label{fig:ThirdItem}
\end{figure}

As expected, we observe a performance drop in the sensory input datasets. In such data, images suffer from potentially high amount of noise (reflections, bad illumination etc.) resulting in noisy visual representations. At the same time, potential cluttering of objects in real-life scenarios also degrades the quality of the proposal generation. These limitations grant our model practically unreliable without any fine-tuning, as we observe by the poor zero-shot performance (below $50\%$). However, the few-shotting experiments suggest a significant performance boost even with minimal fine-tuning. Moreover, we observe that due to its single-stage nature, our model is capable of unravelling ambiguities of reference in the input phrase. Specifically, there are non-sensory examples where the model responds to use of plural, use of definitive pronouns, as well as visual cues, e.g. colour (see Fig.~\ref{fig:ThirdItem}).

\section{Conclusion}
\label{Conclusion}
In this work, we employ a state-of-the-art multi-modal deep learning model and develop an HRI agent for real-time visual grounding. The focus of this work is not on improving the zero-shot model but rather exploiting its single-stage nature, which allows for predictions in unseen data, in order to achieve interaction with humans in dynamic environments (potentially unknown objects) and in natural fashion (robustness to ambiguities of reference). The segmented 3D information of the object can serve as a perceptive utility assisting in navigation, planning, and manipulation actions. We test our methodology by gathering RGB-D data from two robotics-oriented datasets and perform several rounds of experiments to evaluate the systems performance in both zero-shot and few-shot scenarios. 
Experimental results showed that our approach works well in the fine-tuning scenario, both in terms of accuracy and speed. 
In the continuation of this work, we would like to investigate the possibility of ameliorating the low quality of visual representations in sensory images by applying \textit{Synthetic2Real Domain Adaptation}, in effect generating synthetic scenes similar to the setup of our HRI scenario and utilizing them as a fine-tuning resource. 
Another interesting direction is the replacement of the current scene-level visual module with an object-level one, as bounding box extraction can be managed in an RGB-D sensor with the aid of depth data.

\bibliographystyle{abbrv}
\bibliography{revised_version.bib}

\end{document}